\documentclass[10pt, a4paper]{article}
\usepackage{lrec}
\usepackage[dvipsnames]{xcolor}
\usepackage{graphicx}
\usepackage{tabularx}
\usepackage{soul}

\usepackage{epstopdf}
\usepackage[latin1]{inputenc}
\usepackage{comment}
\usepackage{hyperref}
\usepackage{xstring}
\usepackage{amsmath}
\usepackage{booktabs}
\usepackage{cleveref}

\title{Adjusting Image Attributes of Localized Regions with Low-level Dialogue}

\name{Tzu-Hsiang Lin$^{1}$, Alexander Rudnicky$^{1}$, Trung Bui$^{2}$, Doo Soon Kim$^{2}$, Jean Oh$^{1}$}

\address {$^{1}$Carnegie Mellon University, $^{2}$Adobe Research\\
        tzuhsial@alumni.cmu.edu, \{air,jeanoh\}@cmu.edu, \{bui,dkim\}@adobe.com
        }

\abstract{
Natural Language Image Editing~(NLIE) aims to use natural language instructions to edit images.  Since novices are inexperienced with image editing techniques, their instructions are often ambiguous and contain high-level abstractions 
that tend to correspond to complex editing steps to accomplish.
Motivated by this inexperience aspect, we aim to smooth the learning curve by teaching the novices to edit images using low-level commanding terminologies.
Towards this end, we develop a task-oriented dialogue system to investigate low-level instructions for NLIE.  Our system grounds language on the level of edit operations, and suggests options for a user to choose from.  Though compelled to express in low-level terms, a user evaluation shows that 25\% of users found our system easy-to-use, resonating with our motivation.  An analysis shows that users generally adapt to utilizing the proposed low-level language interface. In this study, we identify that object segmentation as the key factor to the user satisfaction. 
Our work demonstrates the advantages of the low-level, direct language-action mapping approach that can be applied to other problem domains beyond image editing such as audio editing or industrial design.   
\\ \newline \Keywords{dialogue, image editing, natural language understanding, low-level} 
}

\begin{document}

\maketitleabstract

\section{Introduction}

Image editing has long been in demand since the invention of photography.  However, learning image editing is time-consuming.  
It involves a wide assortment of features, and combinations of these features to achieve a desired effect.  
As of today, most novices edit their images by requesting experts for help.  
These requests come in the form of Image Edit Requests (IERs) -- natural language descriptions which express desired changes to be made.  
When experts see these requests, they can accurately edit these images via a two-stage process: (i) First, experts look at the original image and the IER and interprets the high-level concepts expressed by the novice. (ii) Second, they come up with into one or more low-level edit operations for these concepts and apply these edits.
Can we build machines that do the same thing? This motivates the study on Natural Language Image Editing (NLIE).

\begin{figure*}
    \centering
    \includegraphics[width=0.6\textwidth]{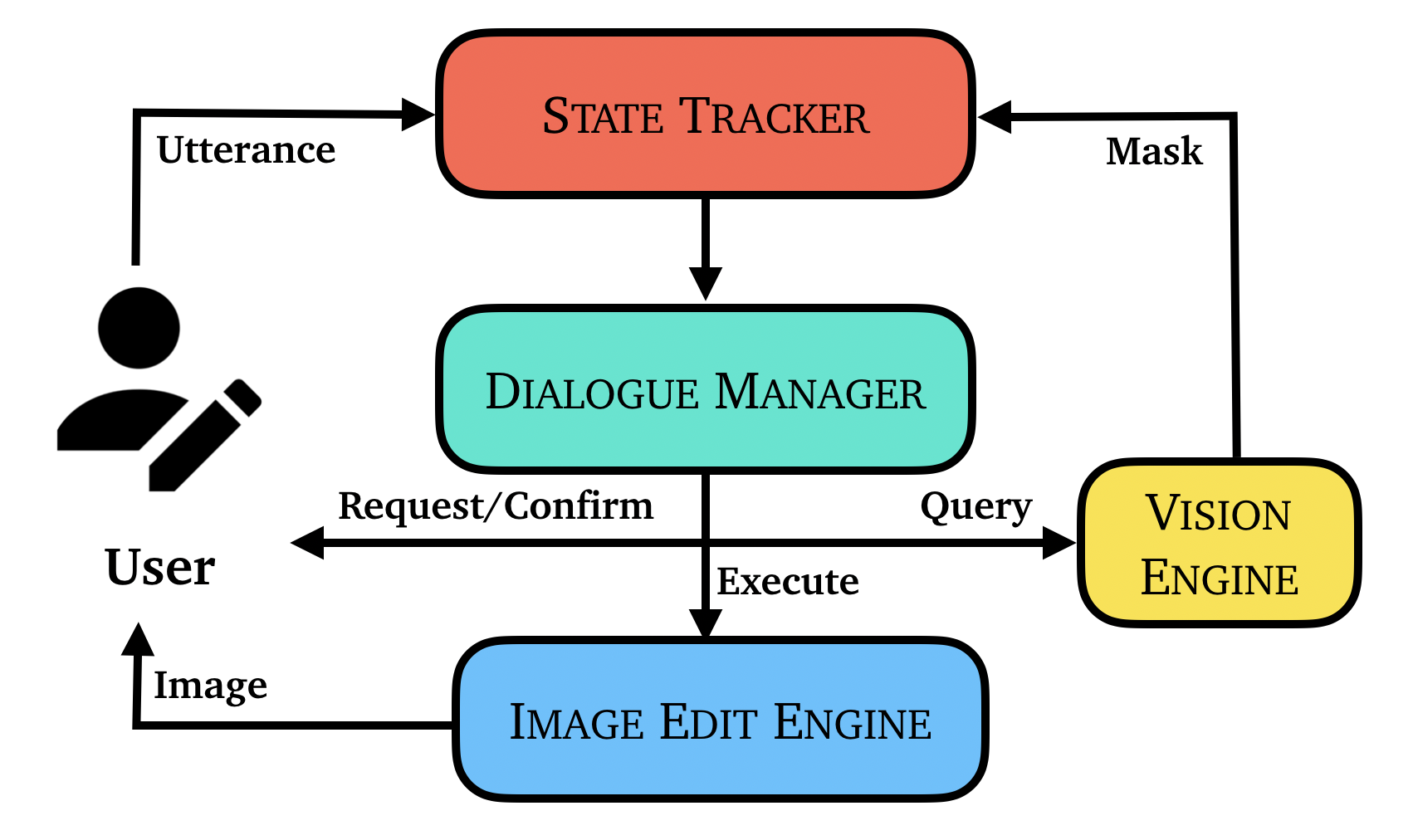}
    \caption{Illustration of our system architecture.  Our system contains 4 components: (i) \textsc{State Tracker}, (ii) \textsc{Dialogue Manager}, (iii) \textsc{Vision Engine}, (iv) \textsc{Image Edit Engine}.  Users interact with \textit{ImaDial} using text inputs.}
    \label{fig:system}
\end{figure*}


There are mainly two approaches towards NLIE.
The two-stage approach follows the editing process of experts.  Several works have developed semantic parsers for the first stage, with datasets collected from crowd-sourcing~\cite{editme}, online image editing communities~\cite{mohapatra2018natural}, or spoken conversation with experts~\cite{cie}.  
Though these semantic parsers are able to capture the high-level intent of novices, little has been discussed about the second stage -- how to infer edits with these parsers.  These datasets still require expert annotations for generating edits, which is a difficult one-to-many mapping.
On the other hand, the image generation approach directly generates an edited image given the original image and an IER using the end-to-end adversarial learning~\cite{goodfellow2014generative} framework.  
This approach is appealing as it learns image edits end-to-end, but also suffers from major drawbacks which limits the number of edit features it can support.
(i) First, image generation is data hungry and domain specific.  To support a new edit feature, a new dataset needs to be collected.  For example, a model trained to modify color of flowers cannot modify color of birds~\cite{chen2018language}. 
(ii) Second, image generation is mostly restricted to global manipulations~\cite{wang2018learning}.  Editing a particular region loses the end-to-end appeal as it requires a separate segmentation module~\cite{lin2014microsoft}, bringing it back to the two-stage approach.

One major challenge we observed in NLIE is the novice language. Having little or no knowledge of terminologies and techniques, novices use open-domain vocabulary to express their needs. As a result, novice IERs have certain characteristics: (i) ambiguous, (ii) abstract (iii) imprecise. Though experts are able to disambiguate or fill-in missing details for novice IERs, grounding open-domain vocabulary have posed difficulties for NLIE.  
For the two-stage approach, annotators have low or near chance level agreement on certain entities~\cite{editme2}, and datasets collected under different scenarios require different annotation schemas~\cite{dialedit}.  
For the image-generation approach, IERs are often imprecise and includes multiple edit operations in a single request~(e.g., enhance white balance and contrast)~\cite{wang2018learning}.  
If novices have some knowledge of image editing tools, machines could understand IERs more easily and will be able to perform edits more precisely.

In this paper we investigate the potential of low-level language for image editing.  
Motivated by the difficulty of open-domain vocabulary, we propose to use dialogue to bridge novice language~(open-domain vocabulary) to image editing terminologies~(in-domain vocabulary).  
Our hypothesis is that, dialogue interaction is an accessible way for users to learn these in-domain low-level terms.
Our end goal is different from previous works as they aim to develop conversational agents~\cite{cie,cheng2018sequential} to complete image edits for users, while we focus on smoothing the learning curve for novices.  

To validate our hypothesis, we developed an image editing dialogue system that grounds on terminologies and conducted a study of 83 users.  Our system is able to segment common objects using referring expressions~\cite{dale1995computational}, and adjust image attributes of these segmented regions.  
Users interact with our system via typed text, and can see the immediate effects of their expressed edits.
Our system received a \textit{Fair} to \textit{Good} rating, and users generally liked the smooth experience of dialogue. 
Through our analysis and user's feedback, we identified object segmentation as the key factor that leads to user (dis-) satisfaction.  We further proposed \textit{Vision Accuracy}, a metric based on dialogue act which correlates with user ratings.

Overall, our contributions can be summarized as follows:
\begin{itemize}
    \item We show that combining dialogue with low-level IERs is a promising solution for supporting NLIE.  
    \item Our analysis and collected dialogues provide insights on user behavior and system aspects of image editing.  Additionally, we propose a novel metric that correlates with user ratings and discuss the strengths and limitations of our system.
    \item We collected two novel datasets to promote future research: (i) a low-level IERs (ii) low-level image editing dialogues. Also, our modular system can be easily extended to support more image edits and be used to collect image editing dialogues.
\end{itemize}

\section{Related Work}

\subsection{Dialogue} 
It is easy to imagine a dialogue for information-seeking domains~\cite{raux2005let,bohus2009ravenclaw,dhingra2017towards,wen2017network,el2017frames,budzianowski2018multiwoz} such as restaurant booking or movie booking, as these conversations exist in everyday life.  Nevertheless, the same could not be said for image editing, as users interact with software tools~(without language input)~\cite{fivek} or post on online communities~(without multi-turn interaction)~\cite{mohapatra2018natural,tan2019expressing}.  
For NLIE, several works have crowdsourced novice IERs by giving image or paired images~\cite{editme,wang2018learning,cheng2018sequential}.  However, they did not consider image editing tools while collecting, and their IERs are often free form in expression.  
The most similar data to our work is the human-to-human dialogues collected in Conversational Image Editing~\cite{cie}, where novice uses only spoken language and an expert carries out the edits.  Our work explores the potential of low-level language for image editing dialogues, focusing on smoothing the learning curve.

\subsection{Language and Vision}  There are many research fields which learns to align semantics between language and visual information, usually with an image and a natural language description.   Our work falls under the category of a more recent setup, where language is related to the difference between paired images.  Several works have been proposed for edited image generation using attention for alignment~\cite{chen2018language,wang2018learning,cheng2018sequential}.   Other works focus more on the language side, either by determining whether a caption is true considering a pair of images~\cite{suhr2019corpus}, or generating captions that describes the difference with latent variables~\cite{jhamtani2018learning} or attention~\cite{tan2019expressing}. Because of using low-level language, our system is simpler as it grounds language at a low-level, and processes visual information with an independent segmentation module.

\begin{figure}[ht]
    \centering
    \includegraphics[width=0.4\textwidth]{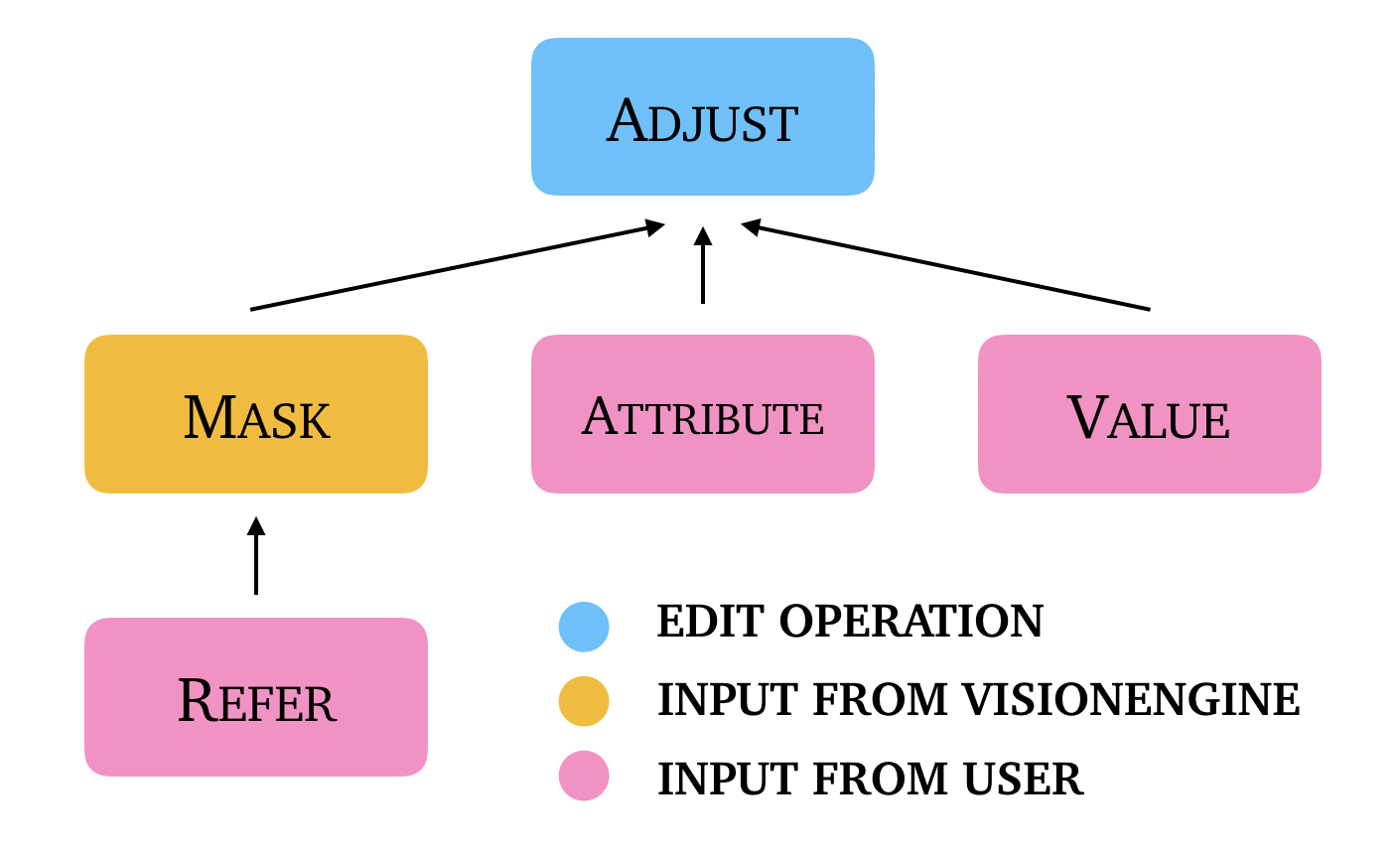}
    \caption{Domain ontology of our system.  \textsc{adjust} operation depends on slots \textsc{mask}, \textsc{attribute}, and \textsc{value}.  Slot \textsc{mask} further depends on slot \textsc{refer}. }
    \label{fig:ontology}
\end{figure}

\section{Task}
\label{sec:task}

\subsection{Definition}
We define the task as instructing the system to perform image edit operations using natural language. To perform this task, the system needs to extract slot values from user's natural language inputs, locate the specified region in the image, and execute the requested edit. 

\subsection{Domain Ontology} 
\label{par:ontology} 

\textsc{Adjust}, which involves manipulating the value of some image property, is the most frequent edit in image editing and is usually realized with a slider; some editing tools provide only \textsc{adjust} operations~\cite{fivek}.  In our system, we study only the \textsc{Adjust} edit, as it already includes image properties which novices have no knowledge about.

To perform \textsc{adjust}, three slot arguments are required~(\Cref{fig:ontology}).  
The first slot is \textsc{mask}, which is a particular region in an image the user wants to modify, and further depends on slot \textsc{refer}. 
This is a referring expression~\cite{krahmer2012computational} given by the user for the system to identify an object in an image.  
The second slot is \textsc{attribute}, which is a modifiable property of an image. The third is \textsc{value}, which is the degree of manipulation of an \textsc{attribute}.  We include 5 common \textsc{attributes}: \{\textit{brightness, contrast, hue, saturation, lightness}\}~\cite{foley1996computer}.  The range of \textsc{value} is from $-100$ to $100$.

\section{Natural Language Understanding}
\label{sec:nlu}

We first describe the key components of our system, which is the Natural Language Understanding (\textsc{NLU}). To build our \textsc{NLU}, we collected low-level IERs~(\Cref{subsec:illcier}) and trained a BIO tagger~(\Cref{subsec:training}) to parse \textsc{refer}, \textsc{attribute}, \textsc{value}.

\subsection{ILLC-IER Dataset}
\label{subsec:illcier}

Image Edit Request (IER) has a broad definition and comes in many different forms. To ensure that the \textsc{NLU} corpus contains instances of all slots in the ontology, we collected sentences that instructs an image edit in imperative form, and includes all the low-level edit arguments.  We refer to these specific type of sentences as ``Imperative Low-Level Complete Image Edit Requests (ILLC-IERs)''.

\paragraph{Collection}  We crowd-sourced ILLC-IERs using the Amazon Mechanical Turk (AMT) platform.  We asked workers to provide a sentence that specifies an image edit and contains \textsc{refer}, \textsc{attribute}, \textsc{value} elements.  For these slots, we provide (i) an image with a highlighted object, (ii) several referring expressions of the highlighted object, (iii) a randomly sampled \textsc{attribute}, and (iv) a randomly sampled \textsc{value}. We used the RefCOCO dataset~\cite{kazemzadeh2014referitgame} for (i) and (ii), and sampled 900 from train, 100 from dev, and 100 from test. 

\paragraph{Annotation} Similar to the data collection, we also crowd-sourced annotations using AMT.  We choose 4 categories for ILLC-IER BIO tagging (i) \textsc{Action} (ii) \textsc{refer} (iii) \textsc{attribute} (iv) \textsc{value}.
\textsc{refer}, \textsc{attribute}, \textsc{value} are slots in our ontology.  \textsc{Action} is the imperative verb that corresponds to the edit action (e.g., ``increase'', ``decrease'', ``modify'') and indicates the sign of \textsc{value}.  For example, ``decrease brightness by 10'' is equivalent to ``change brightness by -10''. 

\paragraph{Statistics}  We collected 2,537 ILLC-IERs.  Based on the data splits in RefCOCO, we split them into 2,055 for train, 242 for dev, and 240 for test. The average number of words per IER is 12.7, and the average length of the referring expressions is 5.4.  The total number of tokens is 32,194, and the number of unique tokens is 1,034.

\subsection{Training BIO Tagger}
\label{subsec:training}

We preprocessed \textsc{attribute} and \textsc{value} by replacing them with abstract tokens \texttt{<attribute>} and \texttt{<value>} since these 2 slots have a closed set of values.
We then trained a one-layer LSTM~\cite{hochreiter1997long} BIO tagger that reached a mean $98.69$ F1 score on the test set.  We also compiled a list of \textsc{action} words that indicate a negative \textsc{value} (e.g., ``decrease'', ``lower'', ``reduce'').

Note that ILLC-IER includes all slot values in a single sentence, which imposes a high degree of constraint, and largely limits the number of ways that users can express IERs.  This establishes our grounding of in-domain vocabulary.

\begin{table*}[]
\centering
\begin{tabular}{c | c c c c | c}
\toprule 
Dataset & \textsc{action} & \textsc{attribute} & \textsc{refer} & \textsc{value} & \textsc{Mean} \\
\midrule
ILLC-IER  &  $96.39$ & $100.0$ & $98.05$ & $100.0$ & $98.69$ \\
Dialogue  &  $63.15$ & $94.59$ & $63.15$ & $99.69$ & $84.13$\\
\bottomrule 
\end{tabular}
\caption{NLU F1 scores on ILLC-IER dataset~(\Cref{sec:nlu}) and user study dialogues~(\Cref{sec:evaluation}).}
\label{tab:nlu}
\end{table*}

\section{System}
\label{sec:system}

Our system architecture is based on the multimodal system of ~\cite{linmultimodal} and consists of 4 components: (i) \textsc{State Tracker}, (ii) \textsc{Dialogue Manager}, (iii) \textsc{Vision Engine}, (iv) \textsc{Image Edit Engine}, along with a web-based interface.   
The main differences between the work of ~\cite{linmultimodal}'s and ours are: (i) their focus is on policy learning with multimodal dialogue state, while we investigate whether users would like the experience of low-level image editing dialogue.  (ii) On the technical side, their system is multimodal which also allows gesture input, while ours only allows text inputs.  However, their \textsc{NLU} is based on regular expressions, while ours is based on recurrent neural networks. \Cref{fig:system} shows the system diagram with dialogue actions.  
We describe each component in detail below.


\subsection{State Tracker}
\label{subsec:tracker}

\textsc{State Tracker} contains the (i) \textsc{NLU} (\Cref{sec:nlu}) component that extracts slot values, and a (ii) \textsc{State Updater} that aggregates \textsc{NLU} outputs over multiple turns.   At every turn, \textsc{State Tracker} takes the current user utterance and the previous turn's dialogue state, and outputs the dialogue state for the current turn.  This dialogue state is then passed to the \textsc{Dialogue Manager}.

\paragraph{NLU} Our \textsc{NLU} is based on the LSTM tagger in \Cref{sec:nlu}, with the addition of a string matcher that detects \{\textit{yes}, \textit{no}\} intents.  The latter handles responses when the \textsc{Dialogue Manager} (\Cref{subsec:manager}) decides to \textit{Confirm} and expects a \textit{yes} or \textit{no} input (e.g., ``Is the current detected region correct? (yes/no)''). 

\paragraph{State Updater} \textsc{State Updater} aggregates turn level outputs from \textsc{NLU}, and clears all slot values when \textsc{Dialogue Manager} decides to \textsc{Execute} an adjustment.  It also removes \textsc{mask} value when a new \textsc{refer} value is detected by NLU.  Our assumption is that all edits are independent, and a new \textsc{refer} should have a new \textsc{mask}.

\subsection{Dialogue Manager}
\label{subsec:manager}

\textsc{Dialogue Manager} takes the current dialogue state (from \textsc{State Tracker}) and outputs an action.  
The actions are: (i) \textit{Request}. (ii) \textit{Confirm} (iii) \textit{Query} (iv) \textit{Execute}. 
\textit{Request} asks the user for a slot value. \textit{Confirm} asks whether the current tracked slot value is correct. \textit{Query} passes the current tracked \textsc{refer} to \textsc{Vision Engine} and receives a \textsc{mask}.  \textit{Execute} sends the tracked \textsc{mask}, \textsc{attribute}, and \textsc{value} to \textsc{Image Edit Engine} for execution.

\paragraph{Policy}  

In image editing,  users will commonly identify a region they want to modify, then decide on the types of edit.  Based on this observation, we use a rule-based policy that obtains slots in the order of \textsc{refer}, \textsc{mask}, \textsc{attribute}, and \textsc{value}.  At the beginning of a dialogue, our \textsc{Dialogue Manager} will first \textit{Request} the slot \textsc{refer}.  After obtaining \textsc{refer}, it will \textit{Query} the \textsc{Vision Engine} then \textit{Confirm} whether the returned \textsc{mask} is correct.  Finally, it will \textit{Request} the slots \textsc{attribute} and \textsc{value} from the user and then \textit{Execute} the \textsc{Adjust} operation.

\paragraph{Suggestive Response}
We use a template-based approach~\cite{raux2005let} to generate system responses.  Since one of our goals is to ease the learning curve, we includes suggestions or options in the templates.
These suggestions aim to guide the user through the image editing process.  For example, we show the value range ``(-100 to 100)'' when asking for \textsc{value}. \Cref{fig:example_bad_vision} shows an example of our policy and suggestive response.

\subsection{Vision Engine} \label{subsec:vision}

The \textsc{Vision Engine} is called when \textsc{Dialogue Manager} decides to \textit{Query}.  The main function of \textsc{Vision Engine} is to infer a localized image region from a user's referring expression.  It acts as an API endpoint for the system, taking image and \textsc{refer} as input and returning a \textsc{mask} that outlines the referred object.  In our system, we return only the \textsc{mask} with the highest confidence score. 

We use MattNet~\cite{yu2018mattnet} as our system's \textsc{Vision Engine}.  MattNet is a modular network that decomposes referring expressions into subject appearance, location, and object relationship components.  It then uses language and visual attention to infer the referred object in the image.  For consistency with \textsc{NLU}, we used MattNet weights trained on RefCOCO (\Cref{sec:nlu}).

\subsection{Image Edit Engine}

The \textsc{Image Edit Engine} displays the image and \textsc{mask} to the user, and performs edits when \textsc{Dialogue Manager} decides to \textit{Execute} an edit operation.  We developed our own Image Edit Engine using OpenCV~\cite{opencv_library}, which is sufficient for our use cases.  Our \textsc{Image Edit Engine} supports the \textsc{attribute} adjustments listed in~\Cref{sec:task}, and can highlight particular regions given a \textsc{mask}. 

\subsection{Interface}

Users interact with our system through a web-based interface.
To simulate an image editing software tool, our interface displays the image, with a highlighted region if \textsc{mask} is provided.  We display the supported list of attributes~(\Cref{sec:task}) and sliders.  We also show the currently tracked \textsc{refer} to users so that they can see the result of their input.
Natural language is the only way to enter IERs;  users are restricted to text inputs (i.e. sliders are not functional).

\begin{table*}[]
    \centering
    \begin{tabular}{c | c c c c c}
    \toprule 
     Rating & \textsc{Very Poor} & \textsc{Poor} & \textsc{Fair} & \textsc{Good} & \textsc{Excellent} \\
    \midrule
      Users      &  $6$ & $22$ & $22$ & $29$ & $4$ \\
    \bottomrule 
    \end{tabular}
    \caption{Performance ratings by users. \textit{ImaDial} received a \textit{Fair} to \textit{Good} rating. }
    \label{tab:performance_rating}
\end{table*}

\section{Evaluation}
\label{sec:evaluation}


\subsection{Setup}

To simulate a real world scenario, we used the test images from our \textsc{NLU} dataset~(\Cref{sec:nlu}), so that both \textsc{NLU} and \textsc{Vision Engine} would be exposed to unseen data.  
We assigned a single AMT user to each image, so each task is done by a different user.\footnote{https://github.com/tzuhsial/ImageEditingWithDialogue}

\paragraph{Instructions} We introduced our system as ``an image editing chatbot that is able to detect objects in the image and adjust several image attributes''.   We asked users to perform at least 2 edits and interact with our system for at least 10 turns. There were no restrictions on what to edit.  To limit overall dialogue length, users could end the dialogue at 30 turns.  Upon task completion, we asked users to rate overall system performance, plus 3 specific system features.  We then asked users to provide feedback on what they liked, disliked and what improvements would make the system better.

\subsection{Results}
\paragraph{Edits}  We assess our system by the number of edits users can complete in each dialogue.  All 83 users completed the required 2 edits, with 5 users exceeded 30 turns. 25 users ($30.1$\%) performed 3 or more edits, with the maximum observed being 5.  This implies that our system is usable, and engaged users enough for them to have done more than the minimum required.

\begin{figure}[t]
    \centering
    \includegraphics[width=0.5\textwidth]{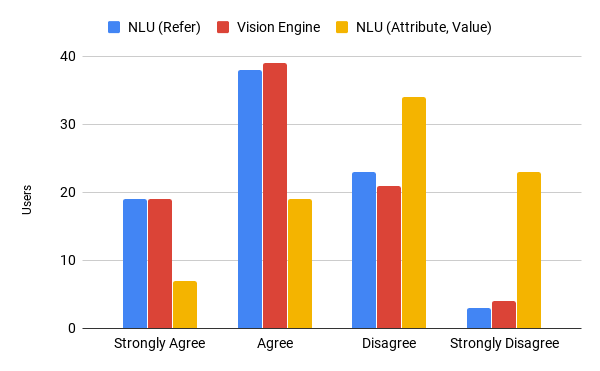}
    \caption{
    Ratings of \textsc{NLU-refer}~(\textcolor{RoyalBlue}{Blue}), \textsc{Vision Engine}~(\textcolor{Maroon}{Red}), and \textsc{NLU-attribute/value}~(\textcolor{YellowOrange}{Yellow}).
    (Strongly) Agree/Disagree were chosen according to a statement ``I found it difficult for the chatbot to use \texttt{<feature>}''. \texttt{<feature>} is the system feature description.
    }
    \label{fig:components_rating}
\end{figure}

\paragraph{Ratings}  Performance and feature ratings are shown in \Cref{tab:performance_rating} and \Cref{fig:components_rating}. Performance ratings are rated on a 5-point Likert Scale from \textit{Very Poor}, \textit{Poor}, \textit{Fair}, \textit{Good}, to \textit{Excellent}. \textit{Good} received the most ratings ($29$), and few users ($10$) rated our system as \textit{Very Poor} or \textit{Excellent}.  
For system features, we choose \textsc{NLU-refer}~(\textsc{NLU} performance on \textsc{refer}), \textsc{Vision Engine}~(Object segmentation), \textsc{NLU-attribute/value}~(\textsc{NLU} performance on \textsc{attribute} and \textsc{value}) for rating.  \textsc{attribute} and \textsc{value} were combined because of similar preprocessing.  
According to a statement ``I found it difficult for the chatbot to \texttt{<feature>}'', where \texttt{<feature>} is a system feature description, system features are rated on a 4-point Likert Scale, from \textit{Strongly Disagree}, \textit{Disagree}, \textit{Agree}, to \textit{Strongly Agree}. Most users agreed the system had difficulty tracking \textsc{refer} and for the \textsc{Vision Engine} inferring the correct \textsc{mask}.  Most users agreed that the system could correctly track  \textsc{attribute} and \textsc{value}.

\section{Analysis and Discussion}
\label{sec:results}

\begin{figure*}
    \centering
    \includegraphics[width=\textwidth]{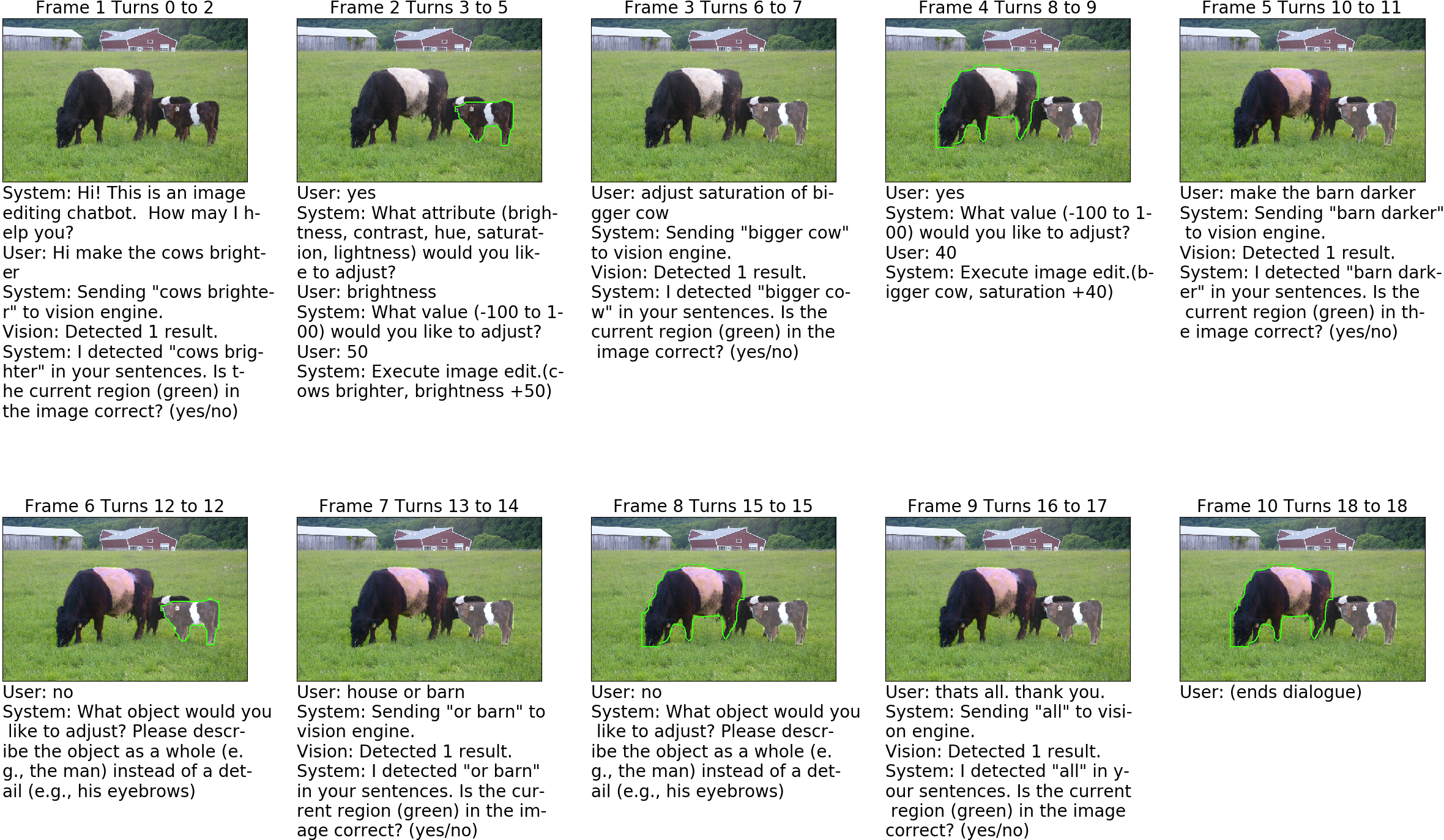}
    \caption{An example dialogue from our user study~(\Cref{sec:evaluation}).  At the beginning of the dialogue, user expressed a high-level IER ``Hi make the cows brighter''.  
    Our system \textit{Confirms} the \textsc{mask} then \textit{Requests} low-level edit arguments \textsc{attribute}, \textsc{value} from the user.  After the 1st edit, user accommodated to our system and expressed an low-level IER ``adjust saturation of bigger cow''~(Frame 3 Turns 6 to 7).  After the 2nd edit, user tried to select the barn ``make the barn darker''~(Frame 5 Turns 10 to 11), and paraphrased ``house or barn''~(Frames 7 Turns 13 to 14) when our \textsc{Vision Engine} failed to detect the correct object.  Finally, user relents and ends the session.}
    \label{fig:example_bad_vision}
\end{figure*}

\paragraph{Statistics} We collected 83 dialogues with 1359 user utterances (2,753 tokens, 534 unique).
The mean number of turns in a dialogue is 17.4; it took an average of 6.3 turns to make an edit. 

\subsection{Quantitative Analysis}

\begin{figure}[t]
    \centering
    \includegraphics[width=0.5\textwidth]{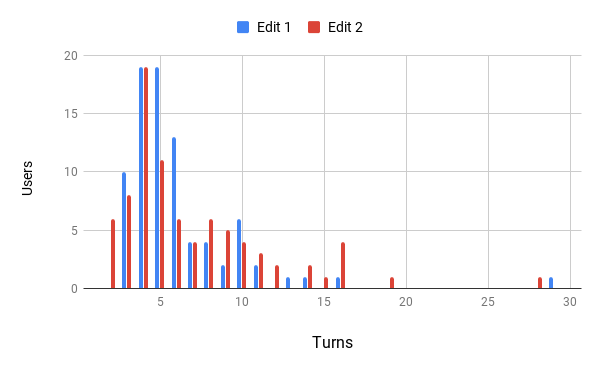}
    \caption{Number of turns to perform the 1st edit (\textcolor{RoyalBlue}{Blue}) and the 2nd edit (\textcolor{Maroon}{Red}).}
    \label{fig:1st_vs_2nd}
\end{figure}

\paragraph{Turns}  Our initial conjecture was that after performing the 1st edit, users would become more familiar with our system and would need fewer of turns.  However, after plotting the turn count the 1st and 2nd edits (\Cref{fig:1st_vs_2nd}), we found that the turn distributions were very similar.  A Kolmogorov-Smirnov test~\cite{massey1951kolmogorov} gives a p-value of $0.93$, indicating sampling from the same distribution.  While this suggests that our system does not have a learning curve, 
this phenomena is likely a result of our \textsc{State Updater}'s clearing all slot values after execution.  Users may want to edit the same object, but have to select \textsc{mask} again.
We leave slot carry-over~\cite{naik2018contextual} for future work.

\paragraph{NLU Coverage} To understand \textsc{NLU} performance in actual dialogue, we annotated dialogue utterances and computed the F1 scores~(\Cref{tab:nlu}).   
Though \textsc{Attribute} and \textsc{value} retained a F1 score of over $90\%$, \textsc{action}, \textsc{refer} dropped from $96\%$ to $63\%$.  This coincides with the system feature ratings from the users, which gave a better rating for \textsc{NLU-attribute/value} than for \textsc{NLU-refer}.

\paragraph{Vision Accuracy}  Object segmentation is important for image editing, and is usually evaluated by Intersection over Union (IoU) with a fixed threshold~\cite{lin2014microsoft}.  Nonetheless, users have many reasons to accept/reject a \textsc{mask}, regardless of IoU.  To assess how \textsc{mask} affects user's ratings, we propose a simple metric to evaluate object segmentation accuracy for image editing systems, based on dialogue actions.

\begin{equation}
\label{eq:visionaccuracy}
\begin{aligned}
    & Vision\ Accuracy =\frac{\#Execute}{\#Query} \\
    & 0 \leq \#Execute \leq \#Query 
    \end{aligned}
\end{equation}

\textit{Vision Accuracy} divides $\#Execute$ (number of \textit{Execute}) by $\#Query$ (number of \textit{Query}).   
Since \textsc{Dialogue Manager} will always \textit{Confirm} \textsc{mask}, an edit (\textit{Execute}) will only contain a \textsc{mask} users deem acceptable.  $\#Query$ can therefore be thought as the number of attempts to get a better \textsc{mask}, with $\#Execute$ being the number of acceptances. \textit{Vision Accuracy} represents the object segmentation performance of the whole system rather than just \textsc{Vision Engine}, since it depends on \textsc{NLU} and our policy.

We computed a Pearson correlation between \textit{Vision Accuracy} and  user ratings.  We found that \textit{Vision Accuracy} has a higher correlation with system performance ($0.46$) than with \textsc{Vision Engine} ratings ($0.32$).
This suggests that our proposed metric, though initially conceived to capture \textsc{Vision Engine} ratings, could also serve as an indicator of system performance.

\subsection{User Feedback}

\paragraph{Likes} After examining user likes, we classified them into 5 categories: (i) Easy to use (ii) Quick (iii) Capable (iv) Experience (v) Other.  For (i), feedback that mentioned the system was easy to use (e.g., ``The edit process seemed straightforward'', ``I liked the ease of use'') or mentioned the system actions \textit{Request} or \textit{Confirm} (e.g., ``it let you know what it wanted'', ``It would ask clarifying questions''). For (ii), we included feedback that mentioned our system was quick.  For (iii),  feedback that mentioned the system was able to understand user utterances, automatically select objects in the image, or perform edits (e.g., ``It understood what I said'', ``Auto select of the area'').  For (iv), feedback that praised the whole experience.  All other feedback went to (v). 

\begin{figure}[t]
    \centering
    \includegraphics[width=0.5\textwidth]{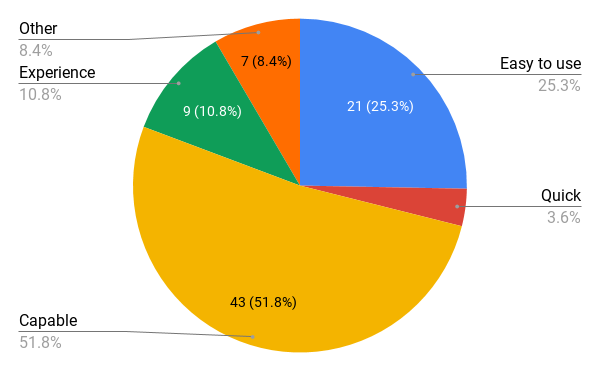}
    \caption{ User feedback on what they liked about our system. We divided into 5 categories (i) Easy to use (\textcolor{RoyalBlue}{Blue}), (ii) Quick (\textcolor{Maroon}{Red}), (iii) Capable (\textcolor{YellowOrange}{Yellow}), (iv) Experience (\textcolor{OliveGreen}{Green}) (v) Other~(\textcolor{Orange}{Orange}).}
    \label{fig:likes}
\end{figure}

This is summarized in \Cref{fig:likes}: users generally liked the concept of an NLIE system (Capable: $51.8$\%), and that dialogue created a smooth experience (Easy to use: $25.3$\%, Experience: $10.8$\%). 

\paragraph{Suggestions}  For dislikes and suggestions, most users first named a thing they disliked, and then suggested improving it.  We report only suggestions from the users.

Most suggestions were centered around object segmentation. 38 users~($46$\%) wanted more accurate \textsc{masks}, with several mentioning that they want the system to show a list of objects it could recognize beforehand.  Several other users suggested including selection tools to modify the predicted \textsc{mask} or just select the object themselves.  

Another trend we observed what users cared about was speed. 9 users~($10.8$\%) suggested using a better \textsc{NLU}, others wanted the system to stop repeating the same mistakes, implying that we should design or learn a better policy~\cite{williams2007partially}. For the interface, some suggested that the system should let users directly modify \textsc{attribute}/\textsc{value} instead text inputs.  One user suggested speech recognition, stating that it would be more efficient to speak and edit at the same time.  Also, another 9 users~($10.8$\%) suggested that responses be made more human-like.

Overall users indicated acceptance if not preference for a language-based interface; other suggestions were for improvements in the language channel. Users seemed to expect that a language interface would somehow be more intelligent and would better understand the task. We believe this is critical requirement for such interfaces. 

\subsection{Discussion}

One of our original goals was to ease the learning curve for image editing tools. From our evaluation and analysis, dialogue actions that guide users through the task (one slot at a time) is an effective way to do this.  The ability to use natural language descriptions for object selection eases overall effort.

The greatest limitation of our current system is that the range of regions that could be selected is bounded by \textsc{Vision Engine} capabilities.  As shown in \Cref{fig:example_bad_vision}, users typically relent after paraphrasing \textsc{refer} multiple times.  Most object segmentation models~\cite{lin2014microsoft} are restricted to detecting common objects, while users may actually want to modify less common objects or non-object regions.  A possible solution is to include selection tools and the option to modify system's inferred \textsc{mask}; region selection is a very delicate procedure.
Users should be able to enjoy the convenience of using referring expressions and fall back to manual selection whenever needed. It could be useful to have a learning system that tries to relate region selection to language.

Apart from object segmentation, another limitation is that users would have to express \textsc{attribute} and \textsc{value} using natural language.  Initially users are unfamiliar with the jargon and the nature of effects; the dialogue manager should suggest one of the possible \textsc{attributes}.  However, as users learn the \textsc{attributes} and how they perform, they would know where the \textsc{attribute} sliders are and preferably manipulate the sliders themselves.  

To summarize, for a task like image editing which exhibits a steep learning curve, dialogue systems can help users get familiar with a complex tool.  However, once users get familiar with the features, they would choose to interact with natural language if it is easier than performing edit operations themselves~\cite{rudnicky1993factors}.  Understanding the characteristics of this threshold should shed more light on the design of dialogue systems for tools that have steep learning curves.  Additionally, it would be interesting to compare against a high-level language based system ~\cite{wang2018learning} and see under which circumstances would the user prefer.  
\footnote{According to the original authors, their data was lost and thus their work could not be compared.}

\section{Conclusion}

Image editing is difficult and takes time to master.  With the goal of improving the learning curve, we developed a low-level task-oriented dialogue system for image adjustments.  Even though our system has only one edit type and requires users to type low-level edit arguments, we find that users take well to a language-based dialogue interface, notable for an essentially visual task. Our work demonstrates the potential of grounding image editing dialogues at low-level, and our analysis shows important aspects for user satisfaction.

For future work, we plan to incorporate additional edit types and develop a multimodal setting that includes gesture manipulations; we expect that language will provide an easy-to-learn interface for users, one that remains available in later use. With the expanded system, we plan to collect a dialogue corpus for dialogue state tracking and training end-to-end models.

\section{Bibliographical References}
\label{main:ref}

\bibliographystyle{lrec}
\bibliography{lrec}


\end{document}